\documentclass[sigconf]{acmart}

\AtBeginDocument{%
  \providecommand\BibTeX{{%
    \normalfont B\kern-0.5em{\scshape i\kern-0.25em b}\kern-0.8em\TeX}}}

\setcopyright{iw3c2w3}
\copyrightyear{2020}
\acmYear{2020}
\acmDOI{10.1145/3366424.3382684}

\acmConference[WWW '20]{Proceedings of The Web Conference 2020}{April 20--24, 2020}{Taipei, Taiwan}
\acmBooktitle{Proceedings of The Web Conference 2020 (WWW '20), April 20--24, 2020, Taipei, Taiwan} 
\acmPrice{}
\acmISBN{978-1-4503-7024-0/20/04}

\begin{document}

\title{TCNN: Triple Convolutional Neural Network Models for Retrieval-based Question Answering System in E-commerce}

\author{Shuangyong Song, Chao Wang, Haiqing Chen, Huan Chen}
\affiliation{%
  \institution{Alibaba Group, Hangzhou 311121, China}
}
\email{{shuangyong.ssy; chaowang.wc; haiqing.chenhq}@alibaba-inc.com, shiwan.ch@taobao.com}








\renewcommand{\shortauthors}{Song, et al.}

\begin{abstract}
  A key solution to the Information Retrieval (IR)-based question answering (QA) models is to retrieve the most similar knowledge entries of a given query from a QA knowledge base, and then rerank those knowledge entries with semantic matching models. In this paper, we aim to improve an IR based e-commerce QA system-AliMe with proposed text matching models, including a basic Triple Convolutional Neural Network (TCNN) model and two Attention-based TCNN (ATCNN) models.
\end{abstract}



\maketitle

\section{Introduction}
In a question-answering knowledge base (KB), each knowledge entry $E$ is stored as a pair: \emph{<knowledge title - $T$, knowledge answer - $A$>}. Most prior IR-based QA systems solely consider semantic relation between a user query $Q$ and a $T$, or semantic relation between a user query and a possible $A$~\cite{QiuLWGCZCHC17}. Both kinds of information can help the detection of final right $A$~\cite{wang2024telechat,shi2025muddit}. So we consider both of them in a unified model to match a user query and an $E$.

Attention Based CNN (ABCNN) model~\cite{DBLP:journals/tacl/YinSXZ16} shows good performance on the task of modeling sentence pairs, since it has a powerful mechanism for modeling two sentences by taking into account the interdependence between them. Referring to the ABCNN models, we propose three ‘Triple Convolutional Neural Network’ (TCNN) models to consider a $Q$, a $T$ and an $A$ together, and two models of them are with attention mechanisms while the other one is not.

\section{Problem Definition and Model Description}

\subsection{Problem Definition}
Given a query and a KB consisting of knowledge entries, we would like to identify entries with strongest semantic relation with the given query. Formally, assume there is a query $Q=(v_1^q,v_2^q,\cdots,v_k^q)$ and there is a knowledge entry $K$ with a title $T=(v_1^t,v_2^t,\cdots,v_m^t)$ and an answer $A=(v_1^a,v_2^a,\cdots,v_n^a)$, where $v_i$ denotes an l-dimensional dense embedding vector of the ith word in the given query, knowledge title or knowledge answer. Our task is to predict the semantic label $y$ (1 for related and 0 for unrelated), which indicates paraphrase identification relation between $Q$ and $K$.

\subsection{Model Description}

\begin{figure}[h]
  \centering
  \includegraphics[width=0.7\linewidth]{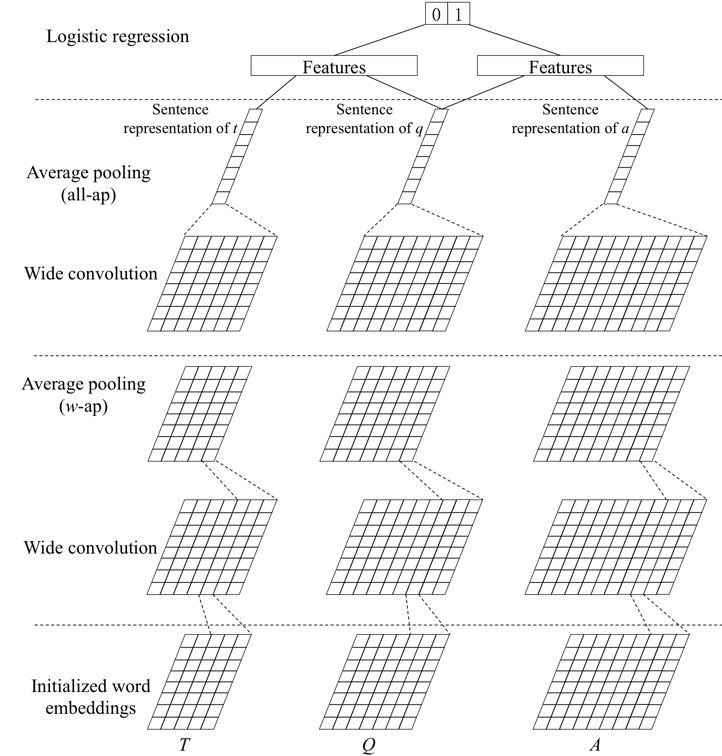}
  \caption{Architecture of TCNN.}
  \label{fig1}
\end{figure}
\textbf{TCNN:} in fig~\ref{fig1}, TCNN consists of three weight-sharing CNNs, each processing one of Q, T and A, and a final layer that solves the query-knowledge matching task.

With convolution layer, we use typical tanh activation function, and with pooling layer, we respectively use average pooling and max pooling as choices. In fig. 1, we just use average pooling as an example.

\begin{figure}[h]
  \centering
  \includegraphics[width=1.0\linewidth]{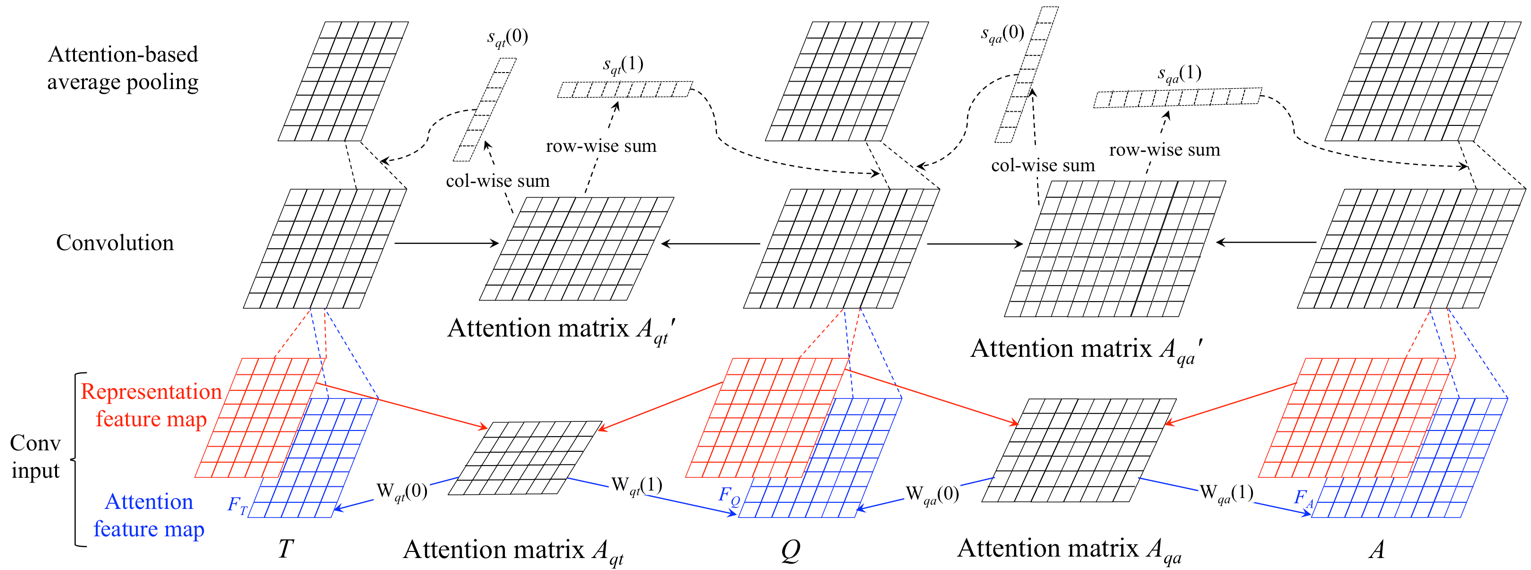}
  \caption{ One block in ATCNN-1.}
  \label{fig2}
\end{figure}

\textbf{ATCNN-1:} Figure 2 shows one block of ATCNN-1, which employs two attention feature matrices Aqt and Aqa to influence convolution. Let Rq be representation feature map of Q and Rt be representation feature map of T, and Aqt(i,j) be the element of Aqt. Then we define Aqt as below, and the acquisition of Aqa is similar.

\begin{equation}
  A_{qt}(i, j) = \cos(R_t[:,i],R_q[:,j])
\end{equation}

With $A_{qa}$ and $A_{qt}$, three attention feature maps $F_T$, $F_Q$ and $F_A$ can be obtained as formula (2). The weight matrices Wqt(0), Wqt(1), Wqa(0) and Wqa(1) are parameters of the model to be learned in training. Different with ABCNN models in which the weights of matrices in this level are shared, we treat these four matrices as different ones and all the weights are trained respectively.

\begin{equation}
{\scriptscriptstyle
  F_T=W_{qt}(0)\cdot (A_{qt})^T, F_Q=\frac{W_{qt}(1)\cdot A_{qt}+W_{qa}(0)\cdot(A_{qa})^T}{2},F_{A}=W_{qa}(1)\cdot A_{qa}}
\end{equation}

On the average pooling layer, $s_{qt}(0)$ is "col-wise sum" of Aqt’ and $s_{qt}(1)$ is "row-wise sum" of Aqt’. $s_{qa}(0)$ and $s_{qa}(1)$ are obtained with same way on Aqa’. Four vectors are used as weights in this pooling step. Particularly, we treat average vector of $s_{qt}(1)$ and $s_{qa}(0)$ as weight vector for the pooling step of $Q$ part.
 
\begin{figure}[h]
  \centering
  \includegraphics[width=1.0\linewidth]{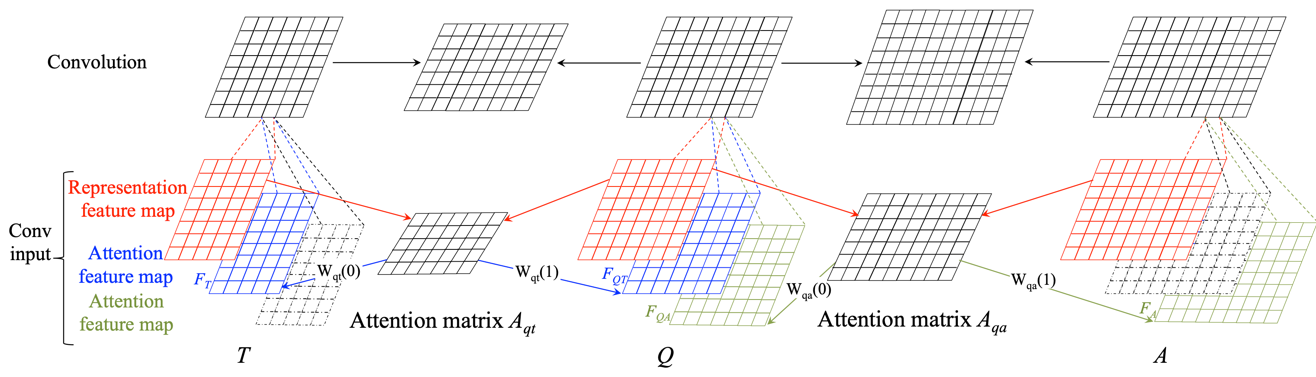}
  \caption{ One block in ATCNN-2.}
  \label{fig2}
\end{figure}

\textbf{ATCNN-2:} Figure 3 shows one block of ATCNN-2. In ATCNN-2, different from ATCNN-1, we respectively transform Aqt and Aqa to two new attention feature maps of Q, as in formula (3), and the input of convolution has three feature maps of Q. To keep three CNNs weight-sharing, we add a zero matrix (black dashed grid) as another new feature map of both T and A. Other parts of ATCNN-2 are the same as of ATCNN-1.
\begin{equation}
  F_{QT}=W_{qt}(1)\cdot A_{qt}, F_{QA}=W_{qa}(0)\cdot (A_{qa})^T
\end{equation}

\section{Experiments}
\textbf{Dataset:} (i) Randomly select 2,000 user queries from chatbot log, and top 15 candidates of each of them can be obtained with Lucene from about 10,000 pieces of <T,A> knowledge, and 8 service experts labeled those candidates with related/unrelated; (ii) Serious data unbalance shows in above labeled data, since just 14.3\% candidates are labeled as related ones. For balancing the dataset, we bring in related ‘query-knowledge’ couples with users’ session-level satisfaction feedbacks. Finally, we got a dataset with 25,423 related couples and 24,034 unrelated couples.

\textbf{Baselines:} (1) BCNN is taken as our first baseline model. (2) ABCNN-3 ~\cite{DBLP:journals/tacl/YinSXZ16}. (3) MatchPyramid is taken as another baseline model. (4) Bi-LSTM. We consider Bi-LSTM as a baseline model and also a representative of RNN based models. (5) WordAverage~\cite{DBLP:conf/emnlp/LiuLSNCP16} is the abbreviation of ‘average of word embeddings’, then we calculate the semantic similarity between two sentences with the Cosine Similarity between their vectors.

\begin{table}[H]
  \caption{Comparison of Different Models’ Performance}
  \label{tab1}
  \begin{tabular}{cllll}
    \toprule
    Methods & Threshold & Precision@1 & Recall@1 & $F_1$@1 \\
    \midrule
    BCNN    & 0.85      & 0.891       & 0.920    & 0.905\\
    ABCNN-3 & 0.63      & 0.878       & 0.953    & 0.914\\
    MatchPyramid & 0.74 & 0.891       & 0.935    & 0.912 \\
    Bi-LSTM & 0.98      & 0.859       & 0.975    & 0.913\\
    WordAverage & 0.93  & 0.618       &\textbf{0.993} &0.762 \\
    TCNN &0.65          &\textbf{0.892} &0.944   &0.918 \\
    ATCNN-1 &0.55       &0.882        &0.969     &0.924 \\
   \textbf{ATCNN-2}&0.46 &0.890       &0.969     &\textbf{0.928} \\
  \bottomrule
\end{tabular}
\end{table}
Table 1 shows threshold-based Precision@1, Recall@1, F1@1 results. Threshold means: if the normalized similarity between two sentences is larger than this threshold, we take them as paraphrase to each other, and vice versa. For all proposed models and baseline models, we choose thresholds with best F1@1.

As expected, TCNN models can get better results than most of baseline models, which shows that knowledge answer content should considered as effective semantic factor in IR-based question answering. For the same reason, ATCNN-2 is better than ABCNN-3. However, ATCNN-1 is not as good as ATCNN-2, and the reason is that ‘relation between query and knowledge title’ is different with ‘relation between query and knowledge answer’, so the acquisition of FQ with combination of Aqt and Aqa is a little unreasonable. In ATCNN-2, we obtain Fqt and Fqa separately for understanding their difference, and the experimental results have shown that it is slightly reasonable than ATCNN-1. Besides, average prediction time of all three TCNN models is just about 1ms, which is fully able to meet the online needs.

\section{Conclusion}

In this paper, we presented TCNN models and demonstrated their effectiveness on QA task. Future work contains modifying TCNN to match user query and knowledge with multiple answers.

\bibliographystyle{ACM-Reference-Format}
\bibliography{sample-base}

\end{document}